\documentclass[10pt,twocolumn,letterpaper]{article}

\usepackage{hhline}
\usepackage{color, colortbl}
\usepackage{multirow}
\usepackage{ijcb}
\usepackage{times}
\usepackage{epsfig}
\usepackage{graphicx}
\usepackage{amsmath}
\usepackage{amssymb}
\usepackage{subcaption}
\usepackage[utf8]{inputenc}
\usepackage{lipsum} 
\usepackage{booktabs}
\usepackage{xcolor}
\usepackage{makecell}
\usepackage{url}

\usepackage[bookmarks=false]{hyperref}

\ijcbfinalcopy 

\ifijcbfinal\fi
\begin{document}

\title{How Does Gender Balance In Training Data Affect Face Recognition Accuracy?}

\author{Vítor Albiero, Kai Zhang and Kevin W. Bowyer\\
University of Notre Dame\\
Notre Dame, Indiana\\
{\tt\small \{valbiero, kzhang4, kwb\}@nd.edu}}

\maketitle
\thispagestyle{empty}

\begin{abstract}
Deep learning methods have greatly increased the accuracy of face recognition, but an old problem still persists: accuracy is usually higher for men than women.
It is often speculated that lower accuracy for women is caused by under-representation in the training data.
This work investigates female under-representation in the training data is truly the cause of lower accuracy for females on test data.
Using a state-of-the-art deep CNN, three different loss functions, and two training datasets, we train each on seven subsets with different male/female ratios, totaling forty two trainings, that are tested on three different datasets\footnote{Trained models are available at \url{https://github.com/vitoralbiero/gender_balance_training_data}}.
Results show that 
(1) gender balance in the training data does not translate into gender balance in the test accuracy, 
(2) the ``gender gap'' in test accuracy is not minimized by a gender-balanced training set, but by a training set with more male images than female images, and
(3) training to minimize the accuracy gap does not result in highest female, male or average accuracy. 
\end{abstract}

\section{Introduction}
\label{introduction}

Deep learning has drastically increased face recognition accuracy
\cite{vgg-face, vggface2, facenet, sphereface, cosface, arcface}.
However, similar to pre-deep-learning face matchers~\cite{frvt, Klare2012, Beveridge2009, grother, grother2010report} 
the accuracy of deep learning methods has usually being shown to be worse for women than men.
(But see a meta-analysis of early research on this topic, which found ambiguous results \cite{Lui2009}.)

Lu et al.~\cite{Lu2018} reported the effects of demographics on unconstrained scenarios using five different deep networks.
Their results show lower accuracy for females, and they speculate that long hair and makeup could be reasons.

In a study of face recognition across ages, Best-Rowden et al.~\cite{longitudinal_c} reported results across several covariates.
In their gender experiment, they report higher match scores across all time intervals for men.
In a extended work~\cite{longitudinal_j}, they added a new dataset, and came to the same conclusion.

Cook et al.~\cite{cook2018} investigate the difference in accuracy between women and men using eleven different automated image acquisition systems.
Using a commercial matcher, their results show higher similarity scores for men than women, suggesting that men are easier to recognize.


NIST recently published a report~\cite{frvt3} focusing in demographic analysis.
On the gender part, they report that women have higher false positive rates than men, and that the phenomenon is consistent across methods and datasets.

\begin{figure}[t]
  \centering
  \includegraphics[width=0.75\columnwidth]{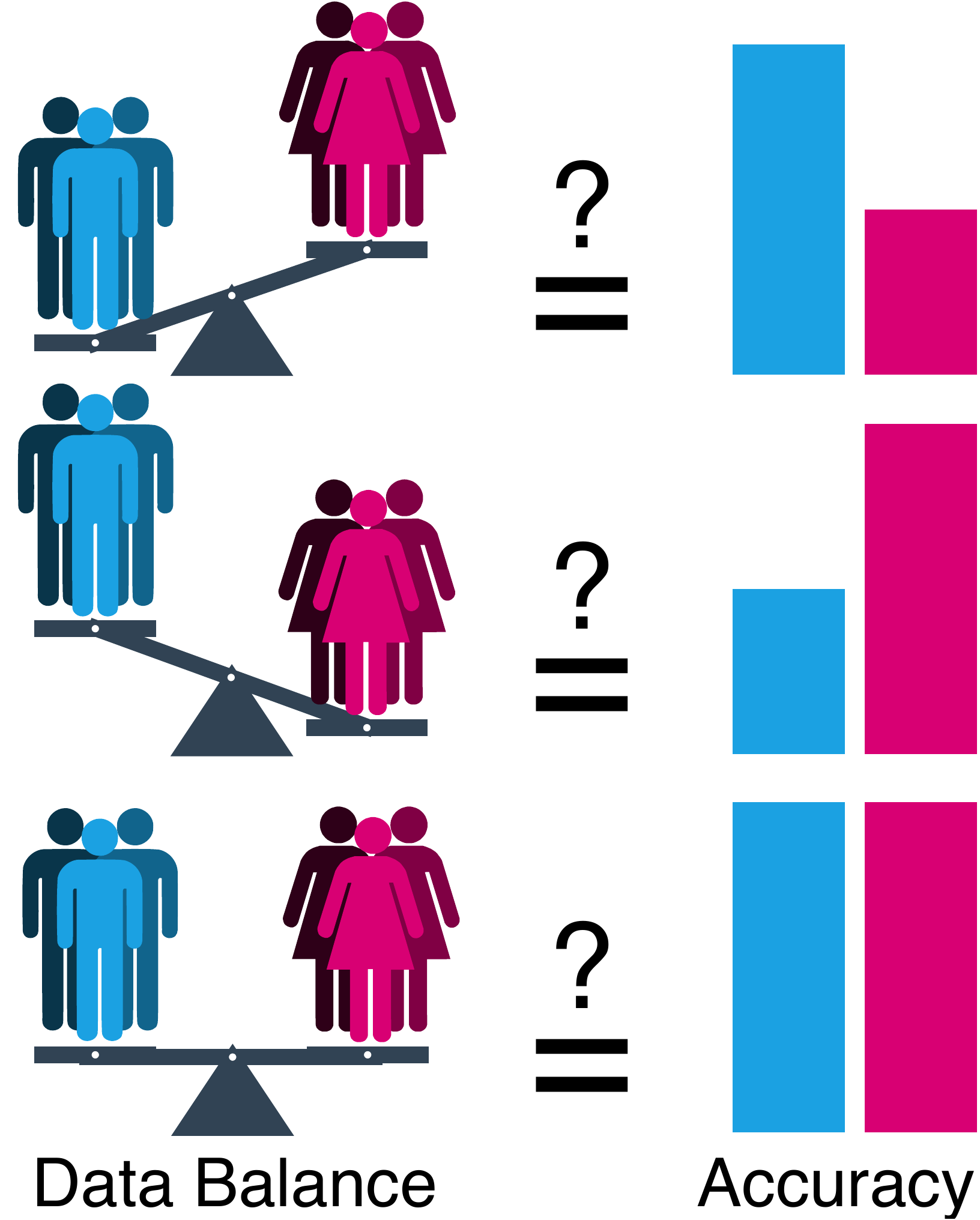}
  \caption{How is the ``gender gap'' in face recognition accuracy related to the gender balance in training data?}
  \label{fig:tease}
  \vspace{-1em}
\end{figure}

Albiero et al.~\cite{albiero_wacvw} report that the separation between authentic and impostor distributions is greater for men than for women. They show that even when images are controlled for (1) makeup, (2) hair covering the 
forehead, (3) head pose and (4) facial expression, the separation is still greater for men.
They also show that the ``gender gap'' in separation of impostor and genuine gets smaller when a balanced training set is used, but men still have a greater separation. 

In deep learning methods, the cause for women having lower accuracy could be under-representation of female images in training~\cite{albiero_wacvw}.
In this work, we investigate how gender distribution of the training data affects accuracy. 
Figure \ref{fig:tease} shows an example of our main hypothesis:
more male data equals better accuracy for males; more female data equals better accuracy for females; and balanced data causes similar accuracy. Using seven training subsets, with different gender-balancing, we investigate how gender balance in training data affects test accuracy.

\begin{figure}[t]
  \centering
  \includegraphics[width=1\columnwidth]{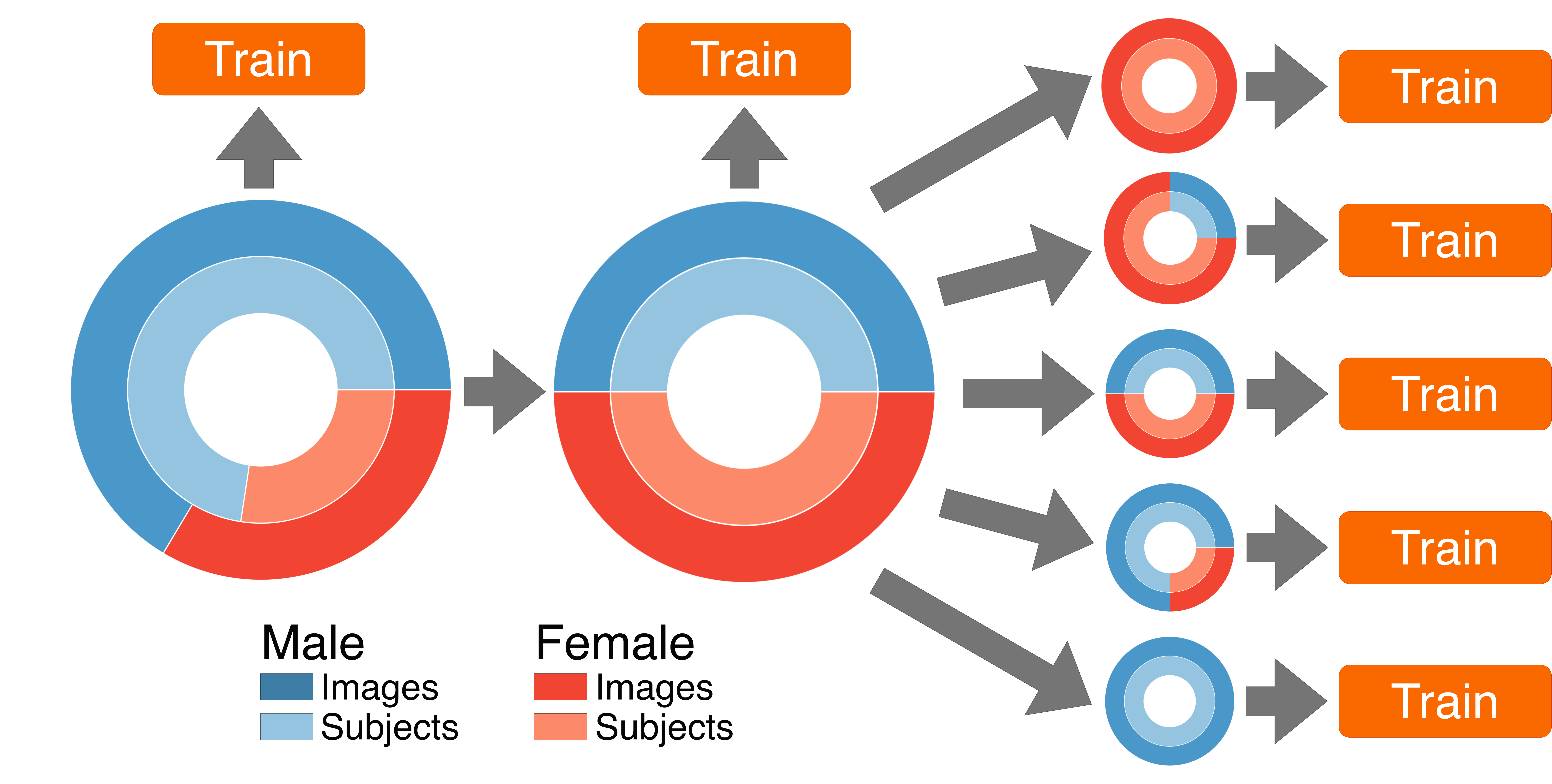}
  \caption{Overview of the proposed training approach.}
  \label{fig:overview}
  \vspace{-1em}
\end{figure}

\section{Method}
This section describes the two training datasets used, how the training subsets were assembled, implementation details for training, and the three testing datasets.
Figure \ref{fig:overview} shows an overview of the proposed training approach.

\subsection{Training Datasets}
\label{training}
We started with two widely-used datasets: VGGFace2~\cite{vggface2} and MS1MV2~\cite{arcface}.
VGGFace2 contains ``wild'', web-scraped images intended to represent a range of pose and age.
VGGFace2 
has 3,477 females (40.3\%) and 5,154 males (59.7\%).
Using the loosely-cropped faces available at~\cite{vggface2_site}, we aligned the faces using the MTCNN~\cite{mtcnn}; 21,025 faces (out of almost 2M) were not detected and these images were dropped.
After alignment, there are 1,291,873 female images (41.4\%) and 1,828,987 male images (58.6\%). 
To create a gender-balanced subset, we randomly removed 1,677 male subjects that account for 537,114 images, to obtain the same number of subjects and images for males and females.
We then assembled five smaller subsets,
with different ratios between males and females.
All the subsets were selected randomly, where a combination containing the desired number of subjects and images was selected.
The top half Table \ref{tab:training_datasets_vggface2} summarizes the subsets assembled from VGGFace2.

\begin{table}[t]
    \centering
    \small
    \begin{tabular}{l|rr|rr}
        \textbf{}& \multicolumn{2}{c|}{\textbf{\# Subjects}}& \multicolumn{2}{c}{\textbf{\# Images}} \\
        \textbf{Subset Name} & \multicolumn{1}{c}{\textbf{Males}} & \multicolumn{1}{c|}{\textbf{Females}} & \multicolumn{1}{c}{\textbf{Males}} & \multicolumn{1}{c}{\textbf{Females}} \\ \hline
        Full & 5,154& 3,477 & 1,828,987& 1,291,873\\
        Balanced & 3,477& 3,477 & 1,291,873& 1,291,873\\ \hline
        F100 & 0& 3,477 & 0& 1,291,873\\
        M25F75 & 870& 2,607 & 322,969& 968,904\\
        M50F50 & 1,739& 1,739 & 645,937& 645,937\\
        M75F25 & 2,607& 870 & 968,904& 322,969\\
        M100 & 3,477& 0 & 1,291,873& 0 \\
        \hline \hline
        Full & 59,563 & 22,499 & 3,741,274 & 1,890,773\\
        Balanced & 22,499 & 22,499 & 1,890,773 & 1,890,773\\ \hline
        F100 & 0 & 22,499 & 0 & 1,890,773\\
        M25F75 & 5,624 & 16,875 & 472,693 & 1,418,080\\
        M50F50 & 11,249 & 11,249 & 945,386 & 945,386\\
        M75F25 & 16,875 & 5,624 & 1,418,080 & 472,693\\
        M100 & 22,499 & 0 & 1,890,773 & 0 
    \end{tabular}
    \vspace{-0.5em}
    \caption{VGGFace2 (top half) and MS1MV2 (bottom half) training subsets created.}
    \vspace{-1em}
    \label{tab:training_datasets_vggface2}
\end{table}

The MS1MV2 dataset~\cite{arcface} is a cleaned version of the MS1M dataset~\cite{ms1_celeb}, containing around 5.8 million images of 85,742 subjects.
We used the MS1MV2 provided at~\cite{insightface}, which already has the faces aligned.
As MS1MV2 does not contain metadata that links back to the original MS1M dataset, we used a gender predictor~\cite{insightface} to label the males and females.
We predicted gender on all the images, and for all subjects that have at least 75\% of its images predicting the same gender, that gender is assigned.
From the 85,742 subjects, 59,563 were assigned as male, and 22,499 were assigned as female.
The 3,680 subjects remaining were removed from the subsets creation, but were used in the full dataset training.
Finally, to create the training subsets, we repeat the same procedure done with VGGFace2 with MS1MV2.
The bottom half of Table \ref{tab:training_datasets_vggface2} summarizes the subsets assembled from MS1MV2.

We train a commonly-used network with three different losses on each one of the subsets, totalling 42 trainings.
As the full and balanced subsets have many more images and subjects than the other five smaller subsets, we analyze their accuracy separately.

\begin{figure*}[t]
    \begin{subfigure}[b]{1\linewidth}
        \centering
        \includegraphics[width=1\columnwidth]{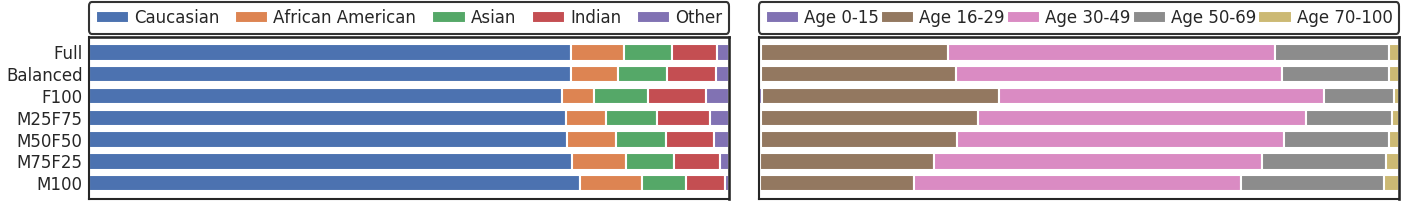}
    \end{subfigure}
    \begin{subfigure}[b]{1\linewidth}
        \vspace{0.25em}
        \centering
        \includegraphics[width=1\columnwidth]{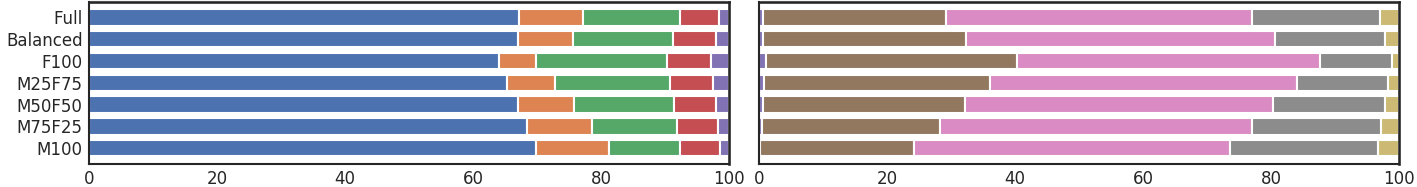}
    \end{subfigure}
    \vspace{-1.25em}
    \caption{Race (left) and age (right) distributions across training subsets for the VGGFace2 (top) and MS1MV2 (bottom) datasets.}
    \label{fig:training_dist}
    \vspace{-1em}
\end{figure*}
\subsubsection{Training Subsets Demographics}
As images were randomly selected, race and age distributions should be similar across the subsets created, as they are a representation of the original dataset, but with different ratios of males and females.
In order to validate if this assumption is correct, we analyze race and age distribution on each subset.
As both the datasets used do not have race or age labels, we train a classifier for each one.
To train the age predictor, we used the AAF~\cite{aaf}, AFAD~\cite{ordinal}, AgeDB~\cite{agedb}, CACD~\cite{cacd}, IMDB-WIKI~\cite{imdb_wiki}, IMFDB~\cite{imfdb}, MegaAgeAsian~\cite{megaageasian}, a commercial version of MORPH~\cite{morph}, and the UTKFace~\cite{utkface} datasets.
The race predictor was trained on the AFAD, IMFDB, MegaAsian, MORPH3, and UTKFace.
For both models, a 90\% and 10\% split was used to train and validate the models.

The race and age models\footnote{Trained models are available at \url{https://github.com/vitoralbiero/face_analysis_pytorch}} were trained using ResNet-50 with modifications as proposed by~\cite{arcface} and~\cite{se}.
The race model was trained using a weighted cross entropy loss, and the age predictor was trained using ordinal regression~\cite{ordinal}.
The race model achieved an accuracy of 97.05\% on the validation set, and the age model a mean absolute error of 4.66. 
Both models and training implementations will be made available.

Figure~\ref{fig:training_dist} shows the race, and age distribution for each subset.
For each subject, the most voted race across its images was assigned.
For better comparison, age was split into 0-15, 16-29, 30-49, 50-69, 70-100 ranges.
The full and balanced subsets for both VGGFace2 and MS1MV2 datasets show near-identical race and age distributions.
For the other smaller subsets, 
the race distributions difference is very small, with the dominant race being Caucasian.
On the other hand, 
going from a females only subset (F100) to a males only subset (M100), we see a decrease in the age 16-29 and an increase in age 50-69.
However, the predominant age group (30-49) is very similar for all the subsets, accounting for approximately 50\% of images.

Finally, given the distributions presented, we believe that race and age are not major confounding factors, thus we can compare the gender factor across each subset.

\subsection{Implementation Details}
\label{training_details}

We train the widely-used ResNet-50~\cite{resnet} architecture using three different loss functions: standard softmax loss, combined margin loss~\cite{arcface}, and triplet loss~\cite{facenet}.
The framework used for training are available at~\cite{insightface}.
We chose these three losses to represent different trainings: a more default loss (softmax); a combination of newer losses (the combined margin loss combines~\cite{cosface, sphereface, arcface}); and a non-classification loss (triplet).
The ResNet-50 used implements modifications suggested by~\cite{arcface} and~\cite{se}, and outputs a feature vector of 512 dimensions.

Stochastic gradient descent (SGD) is used in all the trainings.
The softmax and combined margin loss trainings use mini-batches of 512 images (VGGFace2), and 256 images (MS1MV2).
The triplet loss training uses mini-batches of 240 images for both training datasets, with semi-hard triplet mining and margin of 0.3.
The combined margin training uses the combination M1=1.0, M2=0.3, and M3=0.2, as proposed in the development of ArcFace~\cite{arcface}. 

All trainings are done from scratch, with initial learning rate of 0.1 for the softmax and combined margin losses, and 0.05 for the triplet loss.
For the VGGFace2 trainings, the learning rate is reduced by a factor of 10 at 100K, 140K, and 160K iterations for the bigger subsets (full, balanced) and at 50K, 70K, and 80K for the smaller subsets, and finished at 200K for the larger subsets and 100K for the smaller ones.
For the MS1MV2 trainings, as a smaller mini-batch was used, the trainings ran for twice the number of iterations, finishing at 400K for the bigger subsets, and 200K for the other ones.
The learning rate for the MS1MV2 trainings is reduced at 200K, 280K, and 320K iterations for the two larger subsets, and at 100K, 140K and 160K for the other five.
Face images with size 112x112 are used as input.

We selected the hardest subset of AgeDB~\cite{agedb},
consisting of image pairs that have 30 years difference in subject age,
as a validation set for the trainings.
The AgeDB-30 has around 55\% of its data as males, thus is ``almost'' gender-balanced.
The training is evaluated at every 2K iterations, and the weights with the best TAR@FAR=$0.1\%$ at the AgeDB-30 subset are selected.

\subsection{Test Datasets}
\label{testing_dataset}

\begin{figure}[t]
  \begin{subfigure}[b]{1\linewidth}
      \begin{subfigure}[b]{0.48\linewidth}
        \centering
          \begin{subfigure}[b]{0.4\columnwidth}
            \centering
            \includegraphics[width=\linewidth]{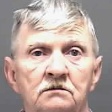}
          \end{subfigure}
          \begin{subfigure}[b]{0.4\columnwidth}
            \centering
            \includegraphics[width=\linewidth]{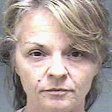}
          \end{subfigure}
          \vspace{-0.25em}
          \caption{MORPH Caucasian}
      \end{subfigure}
      \hfill 
      \begin{subfigure}[b]{0.48\linewidth}
        \centering
          \begin{subfigure}[b]{0.4\columnwidth}
            \centering
            \includegraphics[width=\linewidth]{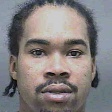}
          \end{subfigure}
          \begin{subfigure}[b]{0.4\columnwidth}
            \centering
            \includegraphics[width=\linewidth]{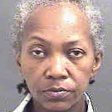}
          \end{subfigure}
          \vspace{-0.25em}
          \caption{MORPH African American}
      \end{subfigure}
      \hfill 
      \begin{subfigure}[b]{0.48\linewidth}
        \centering
          \begin{subfigure}[b]{0.4\columnwidth}
            \centering
            \includegraphics[width=\linewidth]{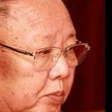}
          \end{subfigure}
          \begin{subfigure}[b]{0.4\columnwidth}
            \centering
            \includegraphics[width=\linewidth]{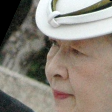}
          \end{subfigure}
          \vspace{-0.25em}
          \caption{IJB-B}
          \vspace{-0.5em}
      \end{subfigure}
      \hfill 
      \begin{subfigure}[b]{0.48\linewidth}
        \centering
          \begin{subfigure}[b]{0.4\columnwidth}
            \centering
            \includegraphics[width=\linewidth]{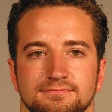}
          \end{subfigure}
          \begin{subfigure}[b]{0.4\columnwidth}
            \centering
            \includegraphics[width=\linewidth]{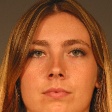}
          \end{subfigure}
          \vspace{-0.25em}
          \caption{Notre Dame}
          \vspace{-0.5em}
      \end{subfigure}
  \end{subfigure}
  \caption{Male and female samples from the test datasets.}
  \label{fig:mean_faces}
  \vspace{-1em}
\end{figure}

We selected three datasets to represent a range of different properties in the test datasets: 
in-the-wild images of highly-varying quality (IJB-B); 
controlled-environment, medium-quality images (MORPH); 
and controlled-environment higher-quality images (Notre Dame).
Figure \ref{fig:mean_faces} shows face samples from the datasets.

MORPH~\cite{morph} 
has been widely used in the research community, and is composed primarily of African Americans and Caucasians, from 16 to 76 years of age.
We used a curated version from~\cite{albiero2019does} that contains a total of 12,775 subjects, of which 10,726 are males and 2,049 are females.
The total number of images is 51,926, and the male and female split is 43,888 and 8,038, respectively.
Since other research~\cite{krishnapria_cvprw_2019} has shown that African-Americans and Caucasians have different accuracy, we analyze the results of the MORPH dataset separately by race.
The African-American subgroup has 41,515 images and the Caucasian subgroup has 10,411.
The faces were detected and aligned as in the training data.

The IARPA Janus B (IJB-B)~\cite{ijbb}
dataset was assembled using celebrity images and videos from the web and is more gender-balanced.
IJB-B contain 991 male subjects (53.72\%) and 854 females subjects (46.28\%).
However, males have 42,989 images (63.03\%) and females only 25,206 images (36.97\%).
Again, faces were aligned using MTCNN.
A total of 2,187 faces were not detected and those images were removed from the experiments.

As the quality of the first two datasets is not ideal, we also assembled a dataset of high-quality images from previously collections at the University of Notre Dame~\cite{frgc}\footnote{The subset image names is available at \url{https://github.com/vitoralbiero/gender_balance_training_data}}.
All images used were acquired in a controlled environment with uniform background.
We removed images that had poor quality, and images that were shot too close to the face.
After curation, we had 261 male subjects with 14,354 images and 169 female subjects with 10,021 images.
The faces were detected and aligned using RetinaFace~\cite{retinaface}.

\section{Experimental Results}
\label{results}
This section presents results for the test datasets: MORPH~\cite{morph}, IJB-B~\cite{ijbb}, and Notre Dame~\cite{frgc}.
We analyze accuracy for the full dataset and the balanced dataset, as well as between the dataset balanced with half the images, and the other four imbalanced ratios of males and females.
The results show in this section are based on false match rates, which should not be affected by the size of the dataset.
Only same-gender images are matched, as cross-gender pairs cannot produce authentic scores, and only contribute to low-similarity impostor scores, thus making the problem easier.
Similarity scores are computed as cosine similarity using the 512 features from the last-but-one layer of the network.

\subsection{Validation}
\label{testing_results}

Validation results are shown in Table \ref{tab:val_results}.
Training using the full and balanced subsets shows very similar results; the largest difference is only 1.01\%.
More interesting, for the five smaller training subsets, for all losses and training datasets, the best accuracy is achieved when the gender-balanced subset (M50F50) is used.
However, the drop-off in accuracy between exact gender-balance (M50F50) and a 75/25 or 25/75 ratio is quite small.
The largest gap in accuracy is seen with the softmax loss and VGGFace2 dataset, where the training using only females (F100) is 5.7\% below the best subset (M50F50).
For the other losses and trainings, the worst accuracy is achieved when training with only females (F100) or only males (M100).
The softmax loss shows overall better accuracy than the other two losses when trained using VGGFace2 subsets.
While training, we observed that models converged earlier when trained with the combined margin loss on some VGGFace2 subsets: the F100 stops improving at only 4k iterations, whereas the M25F75, M75F25, and M100 stopped improving at 10k iterations.
The combined margin loss has better accuracy than the other two losses when trained with the MS1MV2 subsets.
Finally, all three losses show higher accuracy when trained with MS1MV2.

\begin{table}[t]
    \centering
    \small
    \setlength\tabcolsep{5pt}
    \begin{tabular}{l|cc|cc|cc}
\multirow{2}{*}{\textbf{\begin{tabular}[c]{@{}l@{}}Training\\ Subset\end{tabular}}} & \multicolumn{2}{c|}{\textbf{Softmax}} & \multicolumn{2}{c|}{\textbf{Comb. Margin}} & \multicolumn{2}{c}{\textbf{Triplet}} \\
 & \textbf{VGG} & \textbf{MS} & \textbf{VGG} & \textbf{MS} & \textbf{VGG} & \textbf{MS} \\ \hline
Full & \textbf{94.18} & 96.17 & \textbf{94.02} & \textbf{98.03} & 87.23 & 93.65 \\
Balanced & 93.57 & \textbf{96.83} & 93.03 & 97.95 & \textbf{87.43} & \textbf{93.98} \\ \hline
F100 & 86.37 & 93.62 & 85.00 & 94.27 & 81.9 & 87.88 \\
M25F75 & 90.78 & 96.05 & 86.88 & 97.52 & 84.17 & 92.25 \\
M50F50 & \textbf{92.07} & \textbf{96.47} & \textbf{87.97} & \textbf{97.6} & \textbf{84.48} & \textbf{92.93} \\
M75F25 & 91.38 & 96.08 & 87.6 & 97.53 & 84.25 & 92.83 \\
M100 & 87.4 & 93.38 & 85.12 & 95.97 & 78.65 & 87.8
\end{tabular}
    \vspace{-0.5em}
    \caption{Verification rates (\%) on the AgeDB-30 validation set with TAR@FAR=$0.1\%$ for the VGGFace2 (VGG) and MS1MV2 (MS) datasets.}
    \vspace{-1em}
    \label{tab:val_results}
\end{table}
\begin{table*}[]
    \setlength\tabcolsep{4pt}
    \centering
    \small
\begin{tabular}{lll|rrr|rrr|rrr|rrr}
 &  &  & \multicolumn{12}{c}{\textbf{Testing Dataset}} \\
 &  &  & \multicolumn{3}{c|}{\textbf{MORPH C}} & \multicolumn{3}{c|}{\textbf{MORPH AA}} & \multicolumn{3}{c|}{\textbf{IJB-B}} & \multicolumn{3}{c}{\textbf{Notre Dame}} \\
\multirow{-2}{*}{\textbf{Loss}} & \multirow{-2}{*}{\textbf{\begin{tabular}[c]{@{}l@{}}Training\\ Dataset\end{tabular}}} & \multirow{-2}{*}{\textbf{Subset}} & \multicolumn{1}{c}{\textbf{Male}} & \multicolumn{1}{c}{\textbf{Female}} & \multicolumn{1}{l|}{\textbf{Avg.}} & \multicolumn{1}{c}{\textbf{Male}} & \multicolumn{1}{c}{\textbf{Female}} & \multicolumn{1}{l|}{\textbf{Avg.}} & \multicolumn{1}{c}{\textbf{Male}} & \multicolumn{1}{c}{\textbf{Female}} & \multicolumn{1}{l|}{\textbf{Avg.}} & \multicolumn{1}{c}{\textbf{Male}} & \multicolumn{1}{l}{\textbf{Female}} & \multicolumn{1}{l}{\textbf{Avg.}} \\ \hline
 &  & \textbf{Full} & 89.23 & 76.7 & 82.97 & 91.38 & 70.08 & 80.73 & 22.22 & 14.93 & 18.58 & 98.9 & 98.04 & 98.47 \\
 & \multirow{-2}{*}{\textbf{VGGFace2}} & \textbf{Balanced} & \cellcolor[HTML]{86D486}\textbf{90.1} & \cellcolor[HTML]{86D486}\textbf{83.71} & \cellcolor[HTML]{86D486}\textbf{86.91} & \cellcolor[HTML]{86D486}\textbf{94.6} & \cellcolor[HTML]{86D486}\textbf{84.77} & \cellcolor[HTML]{86D486}\textbf{89.69} & \cellcolor[HTML]{86D486}\textbf{33.5} & \cellcolor[HTML]{86D486}\textbf{19.3} & \cellcolor[HTML]{86D486}\textbf{26.4} & \cellcolor[HTML]{86D486}\textbf{99.38} & \cellcolor[HTML]{86D486}\textbf{98.93} & \cellcolor[HTML]{86D486}\textbf{99.16} \\ \cline{2-15} 
 &  & \textbf{Full} & 91.55 & 83.07 & 87.31 & 94.84 & 83.19 & 89.02 & 39.49 & 23.31 & 31.4 & 99.74 & 99.41 & 99.58 \\
\multirow{-4}{*}{\rotatebox[origin=c]{90}{\textbf{Softmax}}} & \multirow{-2}{*}{\textbf{MS1MV2}} & \textbf{Balanced} & \cellcolor[HTML]{86D486}\textbf{95.04} & \cellcolor[HTML]{86D486}\textbf{88.42} & \cellcolor[HTML]{86D486}\textbf{91.73} & \cellcolor[HTML]{86D486}\textbf{97.58} & \cellcolor[HTML]{86D486}\textbf{84.28} & \cellcolor[HTML]{86D486}\textbf{90.93} & \cellcolor[HTML]{86D486}\textbf{40.3} & \cellcolor[HTML]{86D486}\textbf{23.98} & \cellcolor[HTML]{86D486}\textbf{32.14} & \cellcolor[HTML]{86D486}\textbf{99.86} & \cellcolor[HTML]{86D486}\textbf{99.54} & \cellcolor[HTML]{86D486}\textbf{99.7} \\ \hline
 &  & \textbf{Full} & \cellcolor[HTML]{F99C9C}\textbf{93.69} & \cellcolor[HTML]{F99C9C}\textbf{86.44} & \cellcolor[HTML]{F99C9C}\textbf{90.07} & \cellcolor[HTML]{F99C9C}\textbf{95.93} & \cellcolor[HTML]{F99C9C}\textbf{86.68} & \cellcolor[HTML]{F99C9C}\textbf{91.31} & \textbf{0.01} & \textbf{0.01} & \textbf{0.01} & \cellcolor[HTML]{F99C9C}\textbf{99.71} & \cellcolor[HTML]{F99C9C}\textbf{99.43} & \cellcolor[HTML]{F99C9C}\textbf{99.57} \\
 & \multirow{-2}{*}{\textbf{VGGFace2}} & \textbf{Balanced} & 90.38 & 86.29 & 88.34 & 93.31 & 82.76 & 88.04 & \textbf{0.01} & \textbf{0.01} & \textbf{0.01} & 99.33 & 99.23 & 99.28 \\ \cline{2-15} 
 &  & \textbf{Full} & \cellcolor[HTML]{F99C9C}\textbf{99.81} & \cellcolor[HTML]{F99C9C}\textbf{99.39} & \cellcolor[HTML]{F99C9C}\textbf{99.6} & \cellcolor[HTML]{F99C9C}\textbf{99.93} & \cellcolor[HTML]{F99C9C}\textbf{99.7} & \cellcolor[HTML]{F99C9C}\textbf{99.82} & \cellcolor[HTML]{F99C9C}\textbf{52.42} & \cellcolor[HTML]{F99C9C}\textbf{25.67} & \cellcolor[HTML]{F99C9C}\textbf{39.05} & \cellcolor[HTML]{F99C9C}\textbf{100} & 99.97 & 99.99 \\
\multirow{-4}{*}{\rotatebox[origin=c]{90}{\parbox{1.5cm}{\centering\textbf{Combined \\Margin}}}} & \multirow{-2}{*}{\textbf{MS1MV2}} & \textbf{Balanced} & 99.65 & 99.09 & 99.37 & 99.88 & 99.59 & 99.74 & 39.99 & 22.31 & 31.15 & \cellcolor[HTML]{86D486}\textbf{100} & \cellcolor[HTML]{86D486}\textbf{99.99} & \cellcolor[HTML]{86D486}\textbf{100} \\ \hline
 &  & \textbf{Full} & \cellcolor[HTML]{F99C9C}\textbf{78.19} & \cellcolor[HTML]{F99C9C}\textbf{63.67} & \cellcolor[HTML]{F99C9C}\textbf{70.93} & \cellcolor[HTML]{F99C9C}\textbf{82.38} & 59.27 & 70.83 & \cellcolor[HTML]{F99C9C}\textbf{27.45} & 19.32 & \cellcolor[HTML]{F99C9C}\textbf{23.39} & 93.47 & \cellcolor[HTML]{F99C9C}\textbf{91.26} & \cellcolor[HTML]{F99C9C}\textbf{92.37} \\
 & \multirow{-2}{*}{\textbf{VGGFace2}} & \textbf{Balanced} & 77.25 & 62.79 & 70.02 & 81.75 & \cellcolor[HTML]{86D486}\textbf{63.01} & \cellcolor[HTML]{86D486}\textbf{72.38} & 25.95 & \cellcolor[HTML]{86D486}\textbf{19.39} & 22.67 & \cellcolor[HTML]{86D486}\textbf{93.85} & 89.43 & 91.64 \\ \cline{2-15} 
 &  & \textbf{Full} & \cellcolor[HTML]{F99C9C}\textbf{90.68} & 79.56 & 85.12 & \cellcolor[HTML]{F99C9C}\textbf{95.03} & 84.17 & 89.6 & 35.57 & 19.06 & 27.32 & \cellcolor[HTML]{F99C9C}\textbf{98.78} & 97.77 & 98.28 \\
\multirow{-4}{*}{\rotatebox[origin=c]{90}{\textbf{Triplet}}} & \multirow{-2}{*}{\textbf{MS1MV2}} & \textbf{Balanced} & 89.98 & \cellcolor[HTML]{86D486}\textbf{85.1} & \cellcolor[HTML]{86D486}\textbf{87.54} & 94.35 & \cellcolor[HTML]{86D486}\textbf{85.28} & \cellcolor[HTML]{86D486}\textbf{89.82} & \cellcolor[HTML]{86D486}\textbf{35.7} & \cellcolor[HTML]{86D486}\textbf{22.79} & \cellcolor[HTML]{86D486}\textbf{29.25} & 98.36 & \cellcolor[HTML]{86D486}\textbf{99.07} & \cellcolor[HTML]{86D486}\textbf{98.72}
\end{tabular}
\vspace{-0.5em}
\caption{Gender accuracy (\%) with TAR@FAR=$0.001\%$ when trained using the entire training dataset (full) and the gender balanced version (balanced).}
\vspace{-1em}
\label{tab:full_balanced}
\end{table*}

\subsection{Test Accuracy - Full / Balanced Training}

Table~\ref{tab:full_balanced} compares accuracy between training (a) using the full dataset, VGGFace2 or MS1MV2, 
and (b) a subset that is gender-balanced by dropping the required number of male images.
With the softmax loss, training on VGGFace2 or MS1MV2, the gender-balanced training set always results in higher male, female, and average accuracy than the larger, imbalanced full dataset.
Training with softmax almost always results in a lower accuracy than training with combined margin loss.
Training with triplet loss, the accuracy comparison between the full training set and the gender-balanced subset is not very consistent.  A common result, in 4 of 8 instances, is that the gender-balanced training set results in higher female accuracy, higher average accuracy, and lower male accuracy.
Training with triplet loss almost always results in lower accuracy than training with softmax, which in turn is almost always worse than combined margin.
Training with combined margin loss, the full training set results, in 6 of 8 instances, in higher female, male and average accuracy than the gender-balanced training set.  The other two instances are extremes, in that combined margin loss with VGGFace2 fails on IJB-B, and combined margin loss with MS1MV2 on the Notre Dame dataset results in near-perfect accuracy.

The Table also shows the general result that training with MS1MV2 results in higher accuracy than training with VGGFace2.
This, and combined margin loss training with VGGFace2 failing on the IJB-B dataset, are discussed in Section \ref{sec:noise}.


\subsection{Training Data Balance and Testing Accuracy}
\label{sec:hist}
This section analyzes results of five differently gender-balanced training sets, each containing the same number of subjects and images.
Male accuracy is expected to be the best when trained with only male data (M100).
Female accuracy is expected to be the highest when only female data (F100) is used.
The average between males and females is expected to be higher when a balanced subset (M50F50) is used.
Tables shown in this Section are colored from dark green to dark red.
Dark green means the best result is achieved with the expected training, and dark red means the result is achieved with the oppositely balanced training.

\begin{table*}[t]
    \centering
    \small
    \begin{tabular}{llrrrrrrrrrr}
 &  & \multicolumn{10}{c}{\textbf{Training Dataset}} \\
\textbf{} & \multicolumn{1}{r|}{\textbf{}} & \multicolumn{5}{c|}{\textbf{VGGFace2}} & \multicolumn{5}{c}{\textbf{MS1MV2}} \\
\textbf{Loss} & \multicolumn{1}{l|}{} & \multicolumn{1}{c}{\textbf{F100}} & \multicolumn{1}{c}{\textbf{M25F75}} & \multicolumn{1}{c}{\textbf{M50F50}} & \multicolumn{1}{c}{\textbf{M75F25}} & \multicolumn{1}{c|}{\textbf{M100}} & \multicolumn{1}{l}{\textbf{F100}} & \multicolumn{1}{l}{\textbf{M25F75}} & \multicolumn{1}{l}{\textbf{M50F50}} & \multicolumn{1}{l}{\textbf{M75F25}} & \multicolumn{1}{l}{\textbf{M100}} \\ \hline
 & \multicolumn{1}{l|}{\textbf{Male}} & 56.51 & 85 & 86.61 & 89.55 & \multicolumn{1}{r|}{\cellcolor[HTML]{009901}\textbf{89.95}} & 76.85 & 93.08 & 94.51 & 95.21 & \cellcolor[HTML]{009901}\textbf{96.09} \\
 & \multicolumn{1}{l|}{\textbf{Female}} & 80.1 & 80.24 & \cellcolor[HTML]{FFFFC7}\textbf{81.66} & 68.65 & \multicolumn{1}{r|}{55.24} & \cellcolor[HTML]{009901}\textbf{90.52} & 89.41 & 88.74 & 86.34 & 81.47 \\
\multirow{-3}{*}{\textbf{Softmax}} & \multicolumn{1}{l|}{\textbf{Avg.}} & 68.31 & 82.62 & \cellcolor[HTML]{009901}\textbf{84.14} & 79.1 & \multicolumn{1}{r|}{72.6} & 83.69 & 91.25 & \cellcolor[HTML]{009901}\textbf{91.63} & 90.78 & 88.78 \\ \hline
 & \multicolumn{1}{l|}{\textbf{Male}} & 0.41 & 71.25 & 66.95 & 69.53 & \multicolumn{1}{r|}{\cellcolor[HTML]{009901}\textbf{82.98}} & 90.91 & 98.59 & 99.24 & 99.21 & \cellcolor[HTML]{009901}\textbf{99.5} \\
 & \multicolumn{1}{l|}{\textbf{Female}} & 71.13 & \cellcolor[HTML]{86D486}\textbf{71.83} & 57.66 & 63.23 & \multicolumn{1}{r|}{52.64} & \cellcolor[HTML]{009901}\textbf{97.71} & 97.57 & 97.45 & 96.51 & 94.94 \\
\multirow{-3}{*}{\textbf{\begin{tabular}[c]{@{}l@{}}Combined\\ Margin\end{tabular}}} & \multicolumn{1}{l|}{\textbf{Avg.}} & 35.77 & \cellcolor[HTML]{FFFFC7}\textbf{71.54} & 62.31 & 66.38 & \multicolumn{1}{r|}{67.81} & 94.31 & 98.08 & \cellcolor[HTML]{009901}\textbf{98.35} & 97.86 & 97.22 \\ \cline{1-6} \cline{8-12} 
 & \multicolumn{1}{l|}{\textbf{Male}} & 39.52 & 57.23 & 63.88 & 68.37 & \multicolumn{1}{r|}{\cellcolor[HTML]{009901}\textbf{71.36}} & 61.2 & 80.93 & 86.83 & 89.41 & \cellcolor[HTML]{009901}\textbf{90.37} \\
 & \multicolumn{1}{l|}{\textbf{Female}} & 55.96 & \cellcolor[HTML]{86D486}\textbf{56.47} & 53.68 & 53.08 & \multicolumn{1}{r|}{35.24} & 74.61 & 75.84 & \cellcolor[HTML]{FFFFC7}\textbf{77.73} & 76.87 & 67.19 \\
\multirow{-3}{*}{\textbf{Triplet}} & \multicolumn{1}{l|}{\textbf{Avg.}} & 47.74 & 56.85 & 58.78 & \cellcolor[HTML]{FFFFC7}\textbf{60.73} & \multicolumn{1}{r|}{53.3} & 67.91 & 78.39 & \cellcolor[HTML]{009901}\textbf{82.28} & 83.14 & 78.78
\end{tabular}
\vspace{-0.5em}
\caption{Gender accuracy (\%) on the MORPH Caucasian dataset with TAR@FAR=$0.001\%$ with different balancing proportions. All trainings have same number of subjects and images.} 
    \vspace{-1em}
    \label{tab:morph_c}
\end{table*}

Results for the MORPH Caucasian subset are shown in Table \ref{tab:morph_c}.
Except for the M25F75 training using the combined margin loss on VGGFace2, the female accuracy is higher than males only when 100\% female data is used for training (F100).
Training performed on the MS1MV2 dataset shows results that are overall close to what would be expected: the best accuracy on males is when only male data is used; the best accuracy for females is when only female data is used; and the best average accuracy across both is when a gender-balanced training is used.
Moving to VGGFace2, male results still show the same pattern. 
However, females are not as clear.
The highest female accuracy when training with the softmax is with the M50F50 subset, and the highest accuracy with combined margin or triplet is with M25F75.
In the VGGFace2 trainings, the highest average of both groups also disagrees.
With softmax loss it is with the M50F50 subset (expected); with combined margin is using the M25F75 subset; and with triplet loss is achieved using the M75F25 subset.

\begin{table*}[t]
    \centering
    \small
\begin{tabular}{llrrrrrrrrrr}
 &  & \multicolumn{10}{c}{\textbf{Training Dataset}} \\
\textbf{} & \multicolumn{1}{r|}{\textbf{}} & \multicolumn{5}{c|}{\textbf{VGGFace2}} & \multicolumn{5}{c}{\textbf{MS1MV2}} \\
\textbf{Loss} & \multicolumn{1}{l|}{} & \multicolumn{1}{c}{\textbf{F100}} & \multicolumn{1}{c}{\textbf{M25F75}} & \multicolumn{1}{c}{\textbf{M50F50}} & \multicolumn{1}{c}{\textbf{M75F25}} & \multicolumn{1}{c|}{\textbf{M100}} & \multicolumn{1}{l}{\textbf{F100}} & \multicolumn{1}{l}{\textbf{M25F75}} & \multicolumn{1}{l}{\textbf{M50F50}} & \multicolumn{1}{l}{\textbf{M75F25}} & \multicolumn{1}{l}{\textbf{M100}} \\ \hline
 & \multicolumn{1}{l|}{\textbf{Male}} & 55.27 & 87.55 & 90.83 & 89.11 & \multicolumn{1}{r|}{\cellcolor[HTML]{009901}\textbf{92.92}} & 79.19 & 96.28 & 97.58 & 97.95 & \cellcolor[HTML]{009901}\textbf{98.4} \\
 & \multicolumn{1}{l|}{\textbf{Female}} & 63.2 & \cellcolor[HTML]{86D486}\textbf{77.62} & 77.61 & 68.65 & \multicolumn{1}{r|}{58.51} & 88.7 & \cellcolor[HTML]{86D486}\textbf{91.15} & 89.63 & 87.79 & 84.56 \\
\multirow{-3}{*}{\textbf{Softmax}} & \multicolumn{1}{l|}{\textbf{Avg.}} & 59.24 & 82.59 & \cellcolor[HTML]{009901}\textbf{84.22} & 78.88 & \multicolumn{1}{r|}{75.72} & 83.95 & \cellcolor[HTML]{FFFFC7}\textbf{93.72} & 93.61 & 92.87 & 91.48 \\ \hline
 & \multicolumn{1}{l|}{\textbf{Male}} & 72.87 & 76.76 & 62.7 & 70.28 & \multicolumn{1}{r|}{\cellcolor[HTML]{009901}\textbf{84.95}} & 94.79 & 99.43 & 99.72 & 99.78 & \cellcolor[HTML]{009901}\textbf{99.81} \\
 & \multicolumn{1}{l|}{\textbf{Female}} & \cellcolor[HTML]{009901}\textbf{78.2} & 62.47 & 45.36 & 48.97 & \multicolumn{1}{r|}{51.49} & 98.35 & \cellcolor[HTML]{86D486}\textbf{98.85} & 98.44 & 98.58 & 97.88 \\
\multirow{-3}{*}{\textbf{\begin{tabular}[c]{@{}l@{}}Combined\\ Margin\end{tabular}}} & \multicolumn{1}{l|}{\textbf{Avg.}} & \cellcolor[HTML]{F99C9C}\textbf{75.54} & 69.62 & 54.03 & 59.63 & \multicolumn{1}{r|}{68.22} & 96.57 & 99.14 & 99.08 & \cellcolor[HTML]{FFFFC7}\textbf{99.18} & 98.85 \\ \hline
 & \multicolumn{1}{l|}{\textbf{Male}} & 41.5 & 58.5 & 69.09 & 73.63 & \multicolumn{1}{r|}{\cellcolor[HTML]{009901}\textbf{76.77}} & 67.4 & 86.31 & 91.93 & 93.32 & \cellcolor[HTML]{009901}\textbf{94.53} \\
 & \multicolumn{1}{l|}{\textbf{Female}} & 40.38 & 44.41 & \cellcolor[HTML]{FFFFC7}\textbf{49.47} & 47.28 & \multicolumn{1}{r|}{42.88} & 71.34 & 77.51 & 79.72 & \cellcolor[HTML]{F99C9C}\textbf{79.75} & 71.66 \\
\multirow{-3}{*}{\textbf{Triplet}} & \multicolumn{1}{l|}{\textbf{Avg.}} & 40.94 & 51.46 & 59.28 & \cellcolor[HTML]{FFFFC7}\textbf{60.46} & \multicolumn{1}{r|}{59.83} & 69.37 & 81.91 & 85.83 & \cellcolor[HTML]{FFFFC7}\textbf{86.54} & 83.1
\end{tabular}
\vspace{-0.5em}
\caption{Gender accuracy (\%) on the MORPH African American dataset with TAR@FAR=$0.001\%$ with different balancing proportions. All trainings have same number of subjects and images.}
    \vspace{-1em}
    \label{tab:morph_aa}
\end{table*}

The MORPH African American subset results appear in Table \ref{tab:morph_aa}.
Again, females only have higher accuracy than males when training data is 100\% female.
Same as with the Caucasian results, all trainings show highest accuracy for males when training with only male data.
The best female results are less clear, as only the combined margin trained on VGGFace2 dataset shows the highest accuracy for females when trained with 100\% female data.
The majority of the results show a better accuracy for females when a small portion of males are in the training (M25F75), especially the softmax training using the VGGFace2 dataset, as the difference between the 100\% female and 25\% male / 75\% female is 14.42\%.
The most surprising result for females here is the triplet loss trained with the MS1MV2 dataset, as the best result is achieved when trained with more male data than female data (M75F25).
\textit{Looking at average accuracy between males and females, only one loss for one dataset achieved the best accuracy with balanced training.}
For the other losses and datasets, the best average accuracy was achieved with imbalanced training.

\begin{table*}[]
    \centering
    \centering
    \small
    \begin{tabular}{llrrrrrrrrrr}
 &  & \multicolumn{10}{c}{\textbf{Training Dataset}} \\
\textbf{} & \multicolumn{1}{r|}{\textbf{}} & \multicolumn{5}{c|}{\textbf{VGGFace2}} & \multicolumn{5}{c}{\textbf{MS1MV2}} \\
\textbf{Loss} & \multicolumn{1}{l|}{} & \multicolumn{1}{c}{\textbf{F100}} & \multicolumn{1}{c}{\textbf{M25F75}} & \multicolumn{1}{c}{\textbf{M50F50}} & \multicolumn{1}{c}{\textbf{M75F25}} & \multicolumn{1}{c|}{\textbf{M100}} & \multicolumn{1}{l}{\textbf{F100}} & \multicolumn{1}{l}{\textbf{M25F75}} & \multicolumn{1}{l}{\textbf{M50F50}} & \multicolumn{1}{l}{\textbf{M75F25}} & \multicolumn{1}{l}{\textbf{M100}} \\ \hline
 & \multicolumn{1}{l|}{\textbf{Male}} & 85.51 & 96.87 & 97.87 & \cellcolor[HTML]{86D486}\textbf{98.73} & \multicolumn{1}{r|}{98.65} & 95.14 & 99.56 & 99.91 & \cellcolor[HTML]{86D486}\textbf{99.93} & 99.92 \\
 & \multicolumn{1}{l|}{\textbf{Female}} & 98.44 & \cellcolor[HTML]{86D486}\textbf{98.67} & 96.86 & 97.44 & \multicolumn{1}{r|}{87.47} & \cellcolor[HTML]{009901}\textbf{99.89} & 99.82 & 99.67 & 99.54 & 98.22 \\
\multirow{-3}{*}{\textbf{Softmax}} & \multicolumn{1}{l|}{\textbf{Avg.}} & 91.98 & 97.77 & 97.37 & \cellcolor[HTML]{FFFFC7}\textbf{98.09} & \multicolumn{1}{r|}{93.06} & 97.52 & 99.69 & \cellcolor[HTML]{009901}\textbf{99.79} & 99.74 & 99.07 \\ \hline
 & \multicolumn{1}{l|}{\textbf{Male}} & 92.24 & 91.29 & 94.2 & 92.4 & \multicolumn{1}{r|}{\cellcolor[HTML]{009901}\textbf{96.77}} & 99.56 & 99.98 & 99.99 & \cellcolor[HTML]{86D486}\textbf{100} & \cellcolor[HTML]{009901}\textbf{100} \\
 & \multicolumn{1}{l|}{\textbf{Female}} & 95.53 & \cellcolor[HTML]{86D486}\textbf{96.47} & 93.6 & 91.68 & \multicolumn{1}{r|}{82.75} & 99.92 & 99.97 & 99.95 & \cellcolor[HTML]{F99C9C}\textbf{99.98} & 99.86 \\
\multirow{-3}{*}{\textbf{\begin{tabular}[c]{@{}l@{}}Combined\\ Margin\end{tabular}}} & \multicolumn{1}{l|}{\textbf{Avg.}} & 93.89 & 93.88 & \cellcolor[HTML]{009901}\textbf{93.9} & 92.04 & \multicolumn{1}{r|}{89.76} & 99.74 & 99.98 & 99.97 & \cellcolor[HTML]{FFFFC7}\textbf{99.99} & 99.93 \\ \hline
 & \multicolumn{1}{l|}{\textbf{Male}} & 68.94 & 81.5 & 87.36 & 89.18 & \multicolumn{1}{r|}{\cellcolor[HTML]{009901}\textbf{91.26}} & 86.61 & 93.2 & 96.82 & 97.94 & \cellcolor[HTML]{009901}\textbf{98.27} \\
 & \multicolumn{1}{l|}{\textbf{Female}} & 82.17 & 82.7 & 86.18 & \cellcolor[HTML]{86D486}\textbf{87.71} & \multicolumn{1}{r|}{76.3} & 97.54 & 97.54 & 97.24 & \cellcolor[HTML]{F99C9C}\textbf{97.74} & 91.44 \\
\multirow{-3}{*}{\textbf{Triplet}} & \multicolumn{1}{l|}{\textbf{Avg.}} & 75.56 & 82.1 & 86.77 & \cellcolor[HTML]{FFFFC7}\textbf{88.45} & \multicolumn{1}{r|}{83.78} & 92.08 & 95.37 & 97.03 & \cellcolor[HTML]{FFFFC7}\textbf{97.84} & 94.86
\end{tabular}
\vspace{-0.5em}
\caption{Gender accuracy (\%) on the Notre Dame dataset with TAR@FAR=$0.001\%$ with different balancing proportions. All trainings have same number of subjects and images.}
    \vspace{-1em}
    \label{tab:nd}
\end{table*}

Moving to the Notre Dame dataset, Table \ref{tab:nd} shows results for the 30 trainings.
First, we can observe that in general all results have higher accuracy than MORPH.
Once more, except for softmax trained with VGGFace2, males show higher accuracy when trained with only male data.
For the females results, the highest accuracy was achieved when 25\% or 75\% of the training data used was composed by males.
The best averages were not always achieved with a perfectly balanced subset, but the difference between the best average and the average with the gender-balanced training are small.

\begin{table*}[]
    \centering
    \small
    \begin{tabular}{llrrrrrrrrrr}
 &  & \multicolumn{10}{c}{\textbf{Training Dataset}} \\
\textbf{} & \multicolumn{1}{r|}{\textbf{}} & \multicolumn{5}{c|}{\textbf{VGGFace2}} & \multicolumn{5}{c}{\textbf{MS1MV2}} \\
\textbf{Loss} & \multicolumn{1}{l|}{} & \multicolumn{1}{c}{\textbf{F100}} & \multicolumn{1}{c}{\textbf{M25F75}} & \multicolumn{1}{c}{\textbf{M50F50}} & \multicolumn{1}{c}{\textbf{M75F25}} & \multicolumn{1}{c|}{\textbf{M100}} & \multicolumn{1}{l}{\textbf{F100}} & \multicolumn{1}{l}{\textbf{M25F75}} & \multicolumn{1}{l}{\textbf{M50F50}} & \multicolumn{1}{l}{\textbf{M75F25}} & \multicolumn{1}{l}{\textbf{M100}} \\ \hline
 & \multicolumn{1}{l|}{\textbf{Male}} & 19.9 & 29.36 & 18.35 & 31.92 & \multicolumn{1}{r|}{\cellcolor[HTML]{009901}\textbf{34.88}} & 24.41 & 37.88 & 41.02 & 40.94 & \cellcolor[HTML]{009901}\textbf{48.35} \\
 & \multicolumn{1}{l|}{\textbf{Female}} & \cellcolor[HTML]{009901}\textbf{18.76} & 14.23 & 7.18 & 17.81 & \multicolumn{1}{r|}{9.97} & 21.2 & 23.41 & \cellcolor[HTML]{FFFFC7}\textbf{24.01} & 22.37 & 17.33 \\
\multirow{-3}{*}{\textbf{Softmax}} & \multicolumn{1}{l|}{\textbf{Avg.}} & 19.33 & 21.8 & 12.77 & \cellcolor[HTML]{FFFFC7}\textbf{24.87} & \multicolumn{1}{r|}{22.43} & 22.81 & 30.65 & 32.52 & 31.66 & \cellcolor[HTML]{F99C9C}\textbf{32.84} \\ \hline
 & \multicolumn{1}{l|}{\textbf{Male}} & 0.02 & 0.01 & 0.01 & 0.01 & \multicolumn{1}{r|}{\textbf{4.85}} & 35.37 & 38.23 & \cellcolor[HTML]{FFFFC7}\textbf{51.74} & 43.38 & 38.79 \\
 & \multicolumn{1}{l|}{\textbf{Female}} & 0.03 & 0.02 & 0.01 & 0.01 & \multicolumn{1}{r|}{\textbf{2.67}} & 19.44 & 24.24 & \cellcolor[HTML]{FFFFC7}\textbf{27.15} & 22.95 & 20.39 \\
\multirow{-3}{*}{\textbf{\begin{tabular}[c]{@{}l@{}}Combined\\ Margin\end{tabular}}} & \multicolumn{1}{l|}{\textbf{Avg.}} & 0.03 & 0.02 & 0.01 & 0.01 & \multicolumn{1}{r|}{\textbf{3.76}} & 27.41 & 31.24 & \cellcolor[HTML]{009901}\textbf{39.45} & 33.17 & 29.59 \\ \hline
 & \multicolumn{1}{l|}{\textbf{Male}} & 14.6 & 19.55 & 23.28 & 26.14 & \multicolumn{1}{r|}{\cellcolor[HTML]{009901}\textbf{28.96}} & 19.22 & 28.54 & 32.13 & 34.42 & \cellcolor[HTML]{009901}\textbf{35} \\
 & \multicolumn{1}{l|}{\textbf{Female}} & \cellcolor[HTML]{009901}\textbf{14.92} & 15.12 & 16.61 & 16.29 & \multicolumn{1}{r|}{14.02} & 19.72 & 19.6 & 18.75 & \cellcolor[HTML]{F99C9C}\textbf{20.78} & 16.16 \\
\multirow{-3}{*}{\textbf{Triplet}} & \multicolumn{1}{l|}{\textbf{Avg.}} & 14.76 & 17.34 & 19.95 & 21.22 & \multicolumn{1}{r|}{\cellcolor[HTML]{F99C9C}\textbf{21.49}} & 19.47 & 24.07 & 25.44 & \cellcolor[HTML]{FFFFC7}\textbf{27.6} & 25.58
\end{tabular}
\vspace{-0.5em}
\caption{Gender accuracy (\%) on the IJB-B dataset with TAR@FAR=$0.001\%$ with different balancing proportions. All trainings have same number of subjects and images.}
\vspace{-1em}
\label{tab:ijbb}
\end{table*}

Table \ref{tab:ijbb} shows results for the IJB-B dataset.
The accuracy is much lower than with other datasets.
The pattern of highest male accuracy occurring with only male training data is also seen here; only one result shows a better accuracy for males when gender-balanced training is used.
Female accuracy shows a not-as-clear pattern.
For some combinations of loss and training data, the best accuracy is achieved when only female data is used.
However, some results show higher accuracy for females when only half or 25\% of the data is female.
The best averages are less centered on balanced subsets than the previous datasets, with best averages being achieved when more or even only male data is used.
The poor accuracy of the combined margin loss trained with the VGGFace2 is discussed on Section \ref{sec:noise}.

\subsection{Training/Testing Noise Issues}
\label{sec:noise}
In this section we investigate why the combined margin loss when trained with VGGFace2 fails on IJB-B as well as on one MORPH Caucasian result.
When trained with the MS1MV2 dataset, the combined margin shows much higher accuracy on IJB-B, as well as when tested on different testing datasets, the combined margin trained with VGGFace2 shows good accuracy.

As both VGGFace2 and MS1MV2 are similar web-scraped datasets, we speculate that the small number of subjects in VGGFace2 is not enough to train the combined margin loss to perform in-the-wild face matching.
To check this speculation, we randomly select the same number of subjects from the MS1MV2 dataset to match the VGGFace2 male and female numbers (3,477 and 5,154).
However, the number of images in this new subset is much smaller, as it only contains 611,043 images.
We repeat the training using this subset called MS1MV2-Small.

\begin{figure}[t]
    \centering
    \begin{subfigure}[b]{0.48\linewidth}
        \centering
        \includegraphics[width=1\columnwidth]{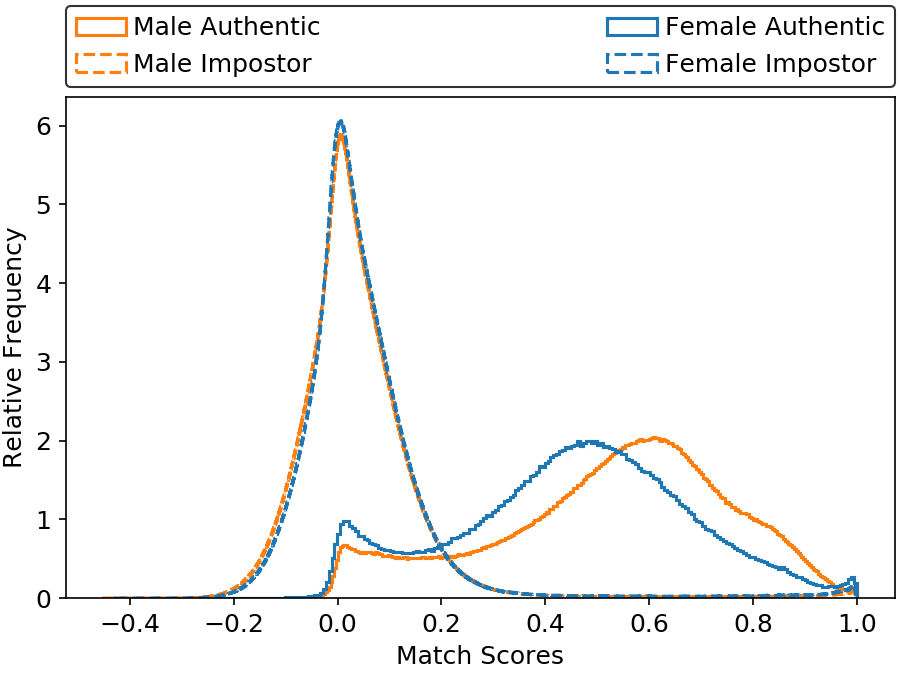}
        \caption{VGGFace2}
        \vspace{-0.5em}
    \end{subfigure}
    \begin{subfigure}[b]{0.48\linewidth}
        \centering
        \includegraphics[width=1\columnwidth]{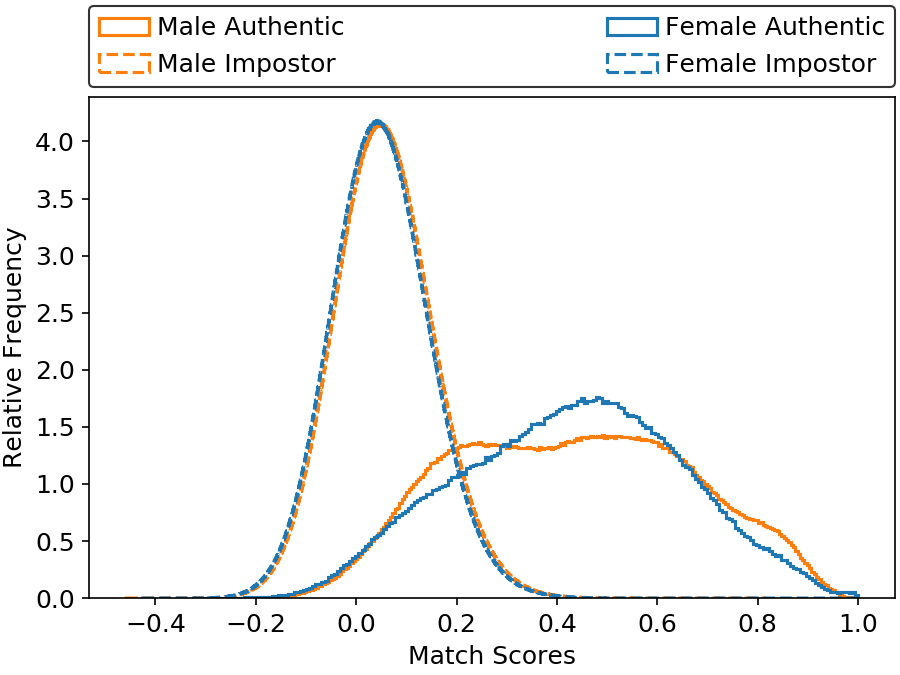}
        \caption{MS1MV2-Small}
        \vspace{-0.5em}
    \end{subfigure}
    \caption{Comparison of authentic and impostor distributions when trained with combined margin loss on VGGFace2 and MS1MV2-Small dataset.}
    \label{fig:vggface2_ms1mv2_small}
    \vspace{-1em}
\end{figure}

Figure \ref{fig:vggface2_ms1mv2_small} shows the authentic and impostor distribution for the MS1MV2-Small compared to the VGGFace2 training.
As the figure shows, the MS1MV2 dataset with same number of subjects has a much higher accuracy, achieving 41.87\% for males and 20.58\% for females with a FAR of $0.001\%$, which clearly shows that the problem is not the number of subjects.
The VGGFace2 test dataset has many mislabeled images, as shown in~\cite{albiero2019does}.
We speculate that the same is true for its training part.
An effort to ``clean'' the dataset to remedy this problem is out of the scope of this paper.

The problem is not only in the training dataset. 
The long tail of the impostor distribution of the VGGFace2 training contains images with low quality, substantial blur and substantial off-frontal pose, which are pushing the match threshold to the end of the authentic distribution.
The same long tail is seen on the MORPH Caucasian result, which is not shown here due to space constraints.
The impostor long tail is much less visible when trained with the MS1MV2 dataset, which was manually cleaned as part of the ArcFace development~\cite{arcface}.

Although the combined margin loss achieves higher accuracy than softmax and triplet loss, it is the only one to be affected by the training noise problems~\cite{wang2018devil} and low quality testing problems~\cite{shi2019probabilistic}.
Together, these cause the matcher to fail catastrophically.
We speculate that, as it is learning margins between subjects, the combined margin is more sensitive to mislabeled data, duplicated subjects, and noise.

\section{Gender-Specific Matcher versus General}
To determine if gender-specific models are better than single models, we compare the average accuracy of males and females when a single model is trained (balanced), and when two specific models are trained (F100 + M100).

\begin{figure}[t]
    \centering
    \includegraphics[width=1\columnwidth]{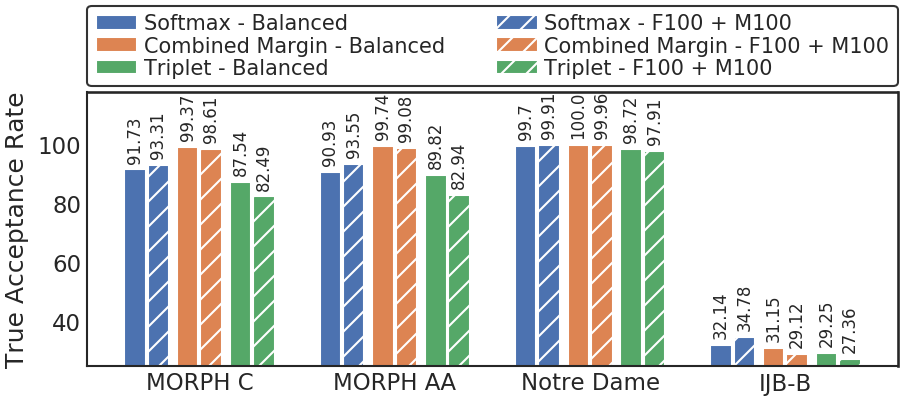}
    \caption{Males and females average TAR@FAR=$0.001\%$ using a single model trained with both genders data, and two models trained with gender specific data.}
    \label{fig:combined_f100_m100}
    \vspace{-1em}
\end{figure}

Figure \ref{fig:combined_f100_m100} shows the comparison between the two approaches.
For the softmax loss, training specific models yields better accuracy on all datasets, with a 2.62\% difference in the MORPH African American subset.
On the other hand, the triplet loss achieves higher accuracy when the single model is trained, with the largest difference in accuracy of 6.88\% on the MORPH African American subset.
Lastly, the combined margin loss shows slightly higher accuracy for the single model, with differences of less than 1\% on the constrained subsets, but 2.13\% on the IJB-B dataset.

\section{Conclusions and Discussion}
\label{conclusions}

All the models and datasets used in this work are available to the research community.
The experiments and results should be reproducible by anyone.

Table~\ref{tab:full_balanced}
compares the accuracy achieved from training with the full version of MS1MV2 or VGGFace2, with the accuracy from training with its gender-balanced subset. 
Note that training with MS1MV2 generally results in higher accuracy than with VGGFace2.
Also note that training with combined margin loss generally results in higher accuracy than with triplet loss, which is generally higher accuracy than softmax.
With softmax, gender-balanced training achieves higher female, male and average accuracy in all instances.
With triplet loss, gender-balanced training achieves higher female and average accuracy in all instances.
However, combined margin loss achieves higher accuracy with the full dataset than with the gender-balanced dataset in 6 of 8 instances.
Thus, a gender-balanced training set, used in combination with a sub-par loss function and training set, generally results in higher female and average accuracy. 
{\it But when training set and loss function are both selected to maximize accuracy, an imbalanced dataset, with more male than female images, results in higher female, male and average accuracy.}

Table~\ref{tab:full_balanced}
includes 24 comparisons of male and female accuracy resulting from explicitly gender-balanced training.  In 22 of the 24, accuracy for males is higher than for females.  Importantly, this includes all four instances for MS1MV2 and combined margin loss.  
{\it Thus, there is little if any empirical support for the premise that training with a gender-balanced training set will result in gender-balanced accuracy on a test set.}

Tables \ref{tab:morph_c} through \ref{tab:ijbb} compare male and female accuracy for equal-sized datasets that vary from 100\% female to 100\% male.
A naïve expectation might be that female accuracy would be maximized with 100\% female training data, male accuracy would be maximized with 100\% male training data, and gender-balanced training would give gender-balanced test accuracy.
However, in all but two instances, gender-balanced training results in higher accuracy for males than females.
Female accuracy is maximized in only 6 of 24 instances with 100\% female training data.
(This finding agrees with~\cite{Klare2012}, which also reported that when an algorithm (non deep learning) was trained with only female data, the female accuracy was worse than when trained with mixed gender data.)
In contrast, male accuracy is maximized in 20 of 24 instances with 100\% male training data.

Focusing on the combined margin loss with the MS1MV2 training set, because this is the highest-accuracy combination, we can make some interesting observations.
Female accuracy is maximized once by training data that is 100\% female, once by training data that is 75\% female, once by training data that is 25\% female, and once by training data that is 50\% female.  Also, the average accuracy is maximized twice by balanced training data and twice by 75\% male training data.
{\it So, again, there is little or no empirical support for the premise that gender-balanced training data will cause gender-balanced test accuracy.}
However, with a good combination of training set and loss function, highest average accuracy may result from training data that is gender-balanced to 75\% male. 

Looking at the gender ratio on the training data that gives the smallest difference between female and male accuracy, it is 25\% male / 75\% female ratio (M25F75) in all four instances for combined margin loss with the MS1MV2 training set.
Thus, it may be possible to choose a gender ratio of the training data to aim for approximately equal test accuracy.
{\it However, target gender ratio for training data that minimizes the gender difference in test accuracy will generally not be 50 / 50.
Also this training will generally not give the highest test accuracy for females, males, or on average.}

The failed results on the IJB-B dataset demonstrate the importance of training dataset curation.
Testing dataset noise is another issue that is observed, which some previous works overcome with ``template'' matching (matching groups of images), instead of single-image matching, but a ``clean'' training dataset is highly desirable.



{\small
\bibliographystyle{ieee}
\bibliography{refs}

\begin{thebibliography}{10}\itemsep=-1pt

\bibitem{vggface2_site}
Vggface2 dataset.
\newblock \url{http://www.robots.ox.ac.uk/~vgg/data/vgg_face2/}.

\bibitem{albiero2019does}
V.~Albiero, K.~W. Bowyer, K.~Vangara, and M.~C. King.
\newblock Does face recognition accuracy get better with age? deep face
  matchers say no.
\newblock In {\em Winter Conference on Applications of Computer Vision (WACV)},
  2020.

\bibitem{albiero_wacvw}
V.~Albiero, K.~K.S., K.~Vangara, K.~Zhang, M.~C. King, and K.~W. Bowyer.
\newblock Analysis of gender inequality in face recognition accuracy.
\newblock In {\em Winter Conference on Applications of Computer Vision
  Workshops (WACVW)}, 2020.

\bibitem{longitudinal_c}
L.~Best-Rowden and A.~K. Jain.
\newblock A longitudinal study of automatic face recognition.
\newblock In {\em International Conference on Biometrics}, 2015.

\bibitem{longitudinal_j}
L.~Best-Rowden and A.~K. Jain.
\newblock Longitudinal study of automatic face recognition.
\newblock {\em IEEE Transactions on Pattern Analysis and Machine Intelligence},
  40(1):148--162, 2018.

\bibitem{Beveridge2009}
J.~R. Beveridge, G.~H. Givens, P.~J. Phillips, and B.~A. Draper.
\newblock {Factors that influence algorithm performance in the Face Recognition
  Grand Challenge}.
\newblock {\em Computer Vision and Image Understanding}, 113(6):750--762, 2009.

\bibitem{vggface2}
Q.~Cao, L.~Shen, W.~Xie, O.~M. Parkhi, and A.~Zisserman.
\newblock Vggface2: A dataset for recognising faces across pose and age.
\newblock In {\em Face and Gesture Recognition}, 2018.

\bibitem{cacd}
B.-C. Chen, C.-S. Chen, and W.~H. Hsu.
\newblock Cross-age reference coding for age-invariant face recognition and
  retrieval.
\newblock In {\em European conference on computer vision}, pages 768--783.
  Springer, 2014.

\bibitem{aaf}
J.~Cheng, Y.~Li, J.~Wang, L.~Yu, and S.~Wang.
\newblock Exploiting effective facial patches for robust gender recognition.
\newblock {\em Tsinghua Science and Technology}, 24(3):333--345, 2019.

\bibitem{cook2018}
C.~M. Cook, J.~J. Howard, Y.~B. Sirotin, and J.~L. Tipton.
\newblock {Fixed and Varying Effects of Demographic Factors on the Performance
  of Eleven Commercial Facial Recognition Systems}.
\newblock {\em IEEE Transactions on Biometrics, Behavior, and Identity
  Science}, 40(1), 2019.

\bibitem{arcface}
J.~Deng, J.~Guo, N.~Xue, and S.~Zafeiriou.
\newblock Arcface: Additive angular margin loss for deep face recognition.
\newblock In {\em Proceedings of the IEEE Conference on Computer Vision and
  Pattern Recognition}, 2019.

\bibitem{retinaface}
J.~Deng, J.~Guo, Y.~Zhou, J.~Yu, I.~Kotsia, and S.~Zafeiriou.
\newblock Retinaface: Single-stage dense face localisation in the wild.
\newblock {\em arXiv preprint arXiv:1905.00641}, 2019.

\bibitem{grother}
P.~Grother.
\newblock Bias in face recognition: What does that even mean? and is it
  serious?
\newblock {\em Biometrics Congress}, 2017.

\bibitem{grother2010report}
P.~J. Grother, G.~W. Quinn, and P.~J. Phillips.
\newblock Report on the evaluation of 2d still-image face recognition
  algorithms.
\newblock 2010.

\bibitem{insightface}
J.~Guo.
\newblock Insightface: 2d and 3d face analysis project.
\newblock https://github.com/deepinsight/insightface, last accessed on November
  2019.

\bibitem{ms1_celeb}
Y.~Guo, L.~Zhang, Y.~Hu, X.~He, and J.~Gao.
\newblock Ms-celeb-1m: A dataset and benchmark for large-scale face
  recognition.
\newblock In {\em European Conference on Computer Vision}, 2016.

\bibitem{resnet}
K.~He, X.~Zhang, S.~Ren, and J.~Sun.
\newblock Deep residual learning for image recognition.
\newblock {\em arXiv preprint arXiv:1512.03385}, 2015.

\bibitem{se}
J.~Hu, L.~Shen, and G.~Sun.
\newblock Squeeze-and-excitation networks.
\newblock In {\em Proceedings of the IEEE conference on computer vision and
  pattern recognition}, pages 7132--7141, 2018.

\bibitem{Klare2012}
B.~F. Klare, M.~J. Burge, J.~C. Klontz, R.~W. {Vorder Bruegge}, and A.~K. Jain.
\newblock {Face recognition performance: Role of demographic information}.
\newblock {\em IEEE Transactions on Information Forensics and Security},
  7(6):1789--1801, 2012.

\bibitem{krishnapria_cvprw_2019}
K.~Krishnapriya, K.~Vangara, M.~C. King, V.~Albiero, and K.~Bowyer.
\newblock Characterizing the variability in face recognition accuracy relative
  to race.
\newblock In {\em Conference on Computer Vision and Pattern Recognition
  Workshops (CVPRW)}, 2019.

\bibitem{sphereface}
W.~Liu, Y.~Wen, Z.~Yu, M.~Li, B.~Raj, and L.~Song.
\newblock Sphereface: Deep hypersphere embedding for face recognition.
\newblock In {\em Proceedings of the IEEE Conference on Computer Vision and
  Pattern Recognition}, 2017.

\bibitem{Lu2018}
B.~Lu, J.~Chen, C.~D. Castillo, and R.~Chellappa.
\newblock An experimental evaluation of covariates effects on unconstrained
  face verification.
\newblock {\em IEEE Transactions on Biometrics, Behavior, and Identity
  Science}, 40(1), 2019.

\bibitem{Lui2009}
Y.~M. Lui, D.~Bolme, B.~A. Draper, J.~R. Beveridge, G.~Givens, and P.~J.
  Phillips.
\newblock {A meta-analysis of face recognition covariates}.
\newblock In {\em IEEE 3rd International Conference on Biometrics: Theory,
  Applications and Systems}, 2009.

\bibitem{agedb}
S.~Moschoglou, A.~Papaioannou, C.~Sagonas, J.~Deng, I.~Kotsia, and
  S.~Zafeiriou.
\newblock Agedb: the first manually collected, in-the-wild age database.
\newblock In {\em Proceedings of the IEEE Conference on Computer Vision and
  Pattern Recognition Workshop}, 2017.

\bibitem{ordinal}
Z.~Niu, M.~Zhou, L.~Wang, X.~Gao, and G.~Hua.
\newblock Ordinal regression with multiple output cnn for age estimation.
\newblock In {\em Proceedings of the IEEE conference on computer vision and
  pattern recognition}, pages 4920--4928, 2016.

\bibitem{vgg-face}
O.~M. Parkhi, A.~Vedaldi, and A.~Zisserman.
\newblock Deep face recognition.
\newblock In {\em BMVC}, 2015.

\bibitem{frvt3}
M.~N. Patrick~Grother and K.~Hanaoka.
\newblock {Face Recognition Vendor Test (FRVT) Part 3: Demographic Effects}.
\newblock {\em NIST IR 8280}, 2003.

\bibitem{frvt}
P.~Phillips, P.~Grother, R.~Micheals, D.~Blackburn, E.~Tabassi, and J.~Bone.
\newblock {Face Recognition Vendor Test 2002: Evaluation Report}.
\newblock {\em NIST IR 6965}, 2003.

\bibitem{frgc}
P.~J. Phillips, P.~J. Flynn, T.~Scruggs, K.~W. Bowyer, J.~Chang, K.~Hoffman,
  J.~Marques, J.~Min, and W.~Worek.
\newblock Overview of the face recognition grand challenge.
\newblock In {\em Computer Vision and Pattern Recognition (CVPR)}, 2005.

\bibitem{morph}
K.~Ricanek and T.~Tesafaye.
\newblock {MORPH: A longitudinal image database of normal adult
  age-progression}.
\newblock In {\em International Conference on Automatic Face and Gesture
  Recognition}, 2006.

\bibitem{imdb_wiki}
R.~Rothe, R.~Timofte, and L.~Van~Gool.
\newblock Deep expectation of real and apparent age from a single image without
  facial landmarks.
\newblock {\em International Journal of Computer Vision}, 126(2-4):144--157,
  2018.

\bibitem{facenet}
F.~Schroff, D.~Kalenichenko, and J.~Philbin.
\newblock Facenet: A unified embedding for face recognition and clustering.
\newblock In {\em IEEE Conference on Computer Vision and Pattern recognition},
  2015.

\bibitem{imfdb}
P.~B. J. G. M. K. R. V. V. H. J. C. K. R. R. R. V.~K. Shankar~Setty,
  Moula~Husain and C.~V. Jawahar.
\newblock {I}ndian {M}ovie {F}ace {D}atabase: {A} {B}enchmark for {F}ace
  {R}ecognition {U}nder {W}ide {V}ariations.
\newblock In {\em National Conference on Computer Vision, Pattern Recognition,
  Image Processing and Graphics (NCVPRIPG)}, Dec 2013.

\bibitem{shi2019probabilistic}
Y.~Shi, A.~K. Jain, and N.~D. Kalka.
\newblock Probabilistic face embeddings.
\newblock {\em arXiv preprint arXiv:1904.09658}, 2019.

\bibitem{wang2018devil}
F.~Wang, L.~Chen, C.~Li, S.~Huang, Y.~Chen, C.~Qian, and C.~Change~Loy.
\newblock The devil of face recognition is in the noise.
\newblock In {\em Proceedings of the European Conference on Computer Vision
  (ECCV)}, pages 765--780, 2018.

\bibitem{cosface}
H.~Wang, Y.~Wang, Z.~Zhou, X.~Ji, D.~Gong, J.~Zhou, Z.~Li, and W.~Liu.
\newblock Cosface: Large margin cosine loss for deep face recognition.
\newblock In {\em Proceedings of the IEEE Conference on Computer Vision and
  Pattern Recognition}, 2018.

\bibitem{ijbb}
C.~Whitelam, E.~Taborsky, A.~Blanton, B.~Maze, J.~Adams, T.~Miller, N.~Kalka,
  A.~K. Jain, J.~A. Duncan, K.~Allen, J.~Cheney, and P.~Grother.
\newblock {IARPA Janus Benchmark-B Face Dataset}.
\newblock {\em IEEE Computer Society Conference on Computer Vision and Pattern
  Recognition Workshops}, 2017-July:592--600, 2017.

\bibitem{mtcnn}
K.~Zhang, Z.~Zhang, Z.~Li, and Y.~Qiao.
\newblock Joint face detection and alignment using multitask cascaded
  convolutional networks.
\newblock {\em IEEE Signal Processing Letters}, 2016.

\bibitem{utkface}
S.~Y. Zhang, Zhifei and H.~Qi.
\newblock Age progression/regression by conditional adversarial autoencoder.
\newblock In {\em IEEE Conference on Computer Vision and Pattern Recognition
  (CVPR)}. IEEE, 2017.

\bibitem{megaageasian}
Y.~Zhang, L.~Liu, C.~Li, et~al.
\newblock Quantifying facial age by posterior of age comparisons.
\newblock 2017.

\end{thebibliography}
}

\end{document}